%% file: neurips_2024.tex
\documentclass[table]{hfstyle/hf}

\usepackage{microtype}
\usepackage{hyperref}
\usepackage{url}
\usepackage{booktabs}
\usepackage{graphicx}
\usepackage{lineno}
\usepackage{enumitem}
\usepackage{listings} %
\usepackage{svg}

\definecolor{ darkblue}{rgb}{0, 0, 0.5}
\hypersetup{colorlinks=true, citecolor=darkblue, linkcolor=darkblue, urlcolor=darkblue}

\usepackage{amssymb}
\usepackage{multirow}
\usepackage{bigdelim}
\usepackage{todonotes}
\usepackage{longtable}
\usepackage{tabularray}
\usepackage{wrapfig}
\usepackage[most]{tcolorbox} 
\usepackage{url}
\usepackage{xspace}
\usepackage{svg}
\usepackage[absolute]{textpos} %

\usepackage{fdsymbol}   %

\usepackage[utf8]{inputenc} %
\usepackage[T1]{fontenc}    %
\usepackage{url}            %
\usepackage{booktabs}       %
\usepackage{amsfonts}       %
\usepackage{nicefrac}       %
\usepackage{microtype}      %
\usepackage[table]{xcolor}         %
\usepackage{amsmath}
\usepackage[most]{tcolorbox}
\usepackage{csquotes}

\usepackage{siunitx}
\usepackage{graphicx}
\usepackage{arydshln}
\usepackage{wrapfig}
\usepackage{enumitem}
\usepackage{soul} %

\usepackage{multirow}
\usepackage{xspace}
\usepackage{adjustbox}
\usepackage{pifont}
\usepackage{caption}
\usepackage{makecell}
\usepackage{bold-extra}
\usepackage{url}

\usepackage{pgf-pie}

\usepackage{hyperref}
\definecolor{linkcolor}{RGB}{0, 0, 128}
\hypersetup{
     colorlinks   = true,
     citecolor    = linkcolor,
     linkcolor    = linkcolor,
     urlcolor     = linkcolor,
}
\usepackage{pifont}%
\usepackage{listings}

\setlist[itemize]{leftmargin=*,itemsep=0em,parsep=0.3em,topsep=0.3em}

\DeclareUnicodeCharacter{2212}{\ensuremath{-}}

\definecolor{maroon}{HTML}{F26035}
\definecolor{yellow}{HTML}{FDBC42}
\definecolor{lavender}{HTML}{734f96}
\definecolor{darkergrey}{HTML}{444444}
\definecolor{midgrey}{HTML}{e6eded}

\definecolor{neutralEight}{HTML}{343434}
\definecolor{neutralFive}{HTML}{838383}
\definecolor{neutralThree}{HTML}{bebebe}
\definecolor{neutralOne}{HTML}{dedede}
\definecolor{lightgrey}{HTML}{fafcfc}

\usepackage{tikz}

\definecolor{maroon}{HTML}{F26035}
\definecolor{yellow}{HTML}{FDBC42}
\definecolor{darkred}{RGB}{156, 39, 33}
\definecolor{darkblue}{RGB}{31, 90, 153}
\definecolor{forestgreen}{rgb}{0.13, 0.55, 0.13}
\definecolor{olmoDarkBlue}{HTML}{012e59}
\definecolor{olmoBlue}{HTML}{265ed4}
\definecolor{olmoLightBlue}{HTML}{012e59}
\definecolor{olmoTeal}{HTML}{00d5ff}
\definecolor{olmoYellow}{HTML}{ffbb00}
\definecolor{olmoOrange}{HTML}{ff9100}

\usepackage{setspace}

\usepackage{nicematrix}
\newcolumntype{L}[1]{>{\raggedright\let\newline\\\arraybackslash\hspace{0pt}}m{#1}}
\newcolumntype{C}[1]{>{\centering\let\newline\\\arraybackslash\hspace{0pt}}m{#1}}
\newcolumntype{R}[1]{>{\raggedleft\let\newline\\\arraybackslash\hspace{0pt}}m{#1}}
\newcolumntype{P}[1]{>{\centering\let\newline\\\arraybackslash\columncolor{ai2lightpink}}m{#1}}
\input{preamble}

\title{SmolVLM: Redefining small and efficient multimodal models}

\newcommand{\huggingface}{\raisebox{-1.5pt}{\includegraphics[height=1.05em]{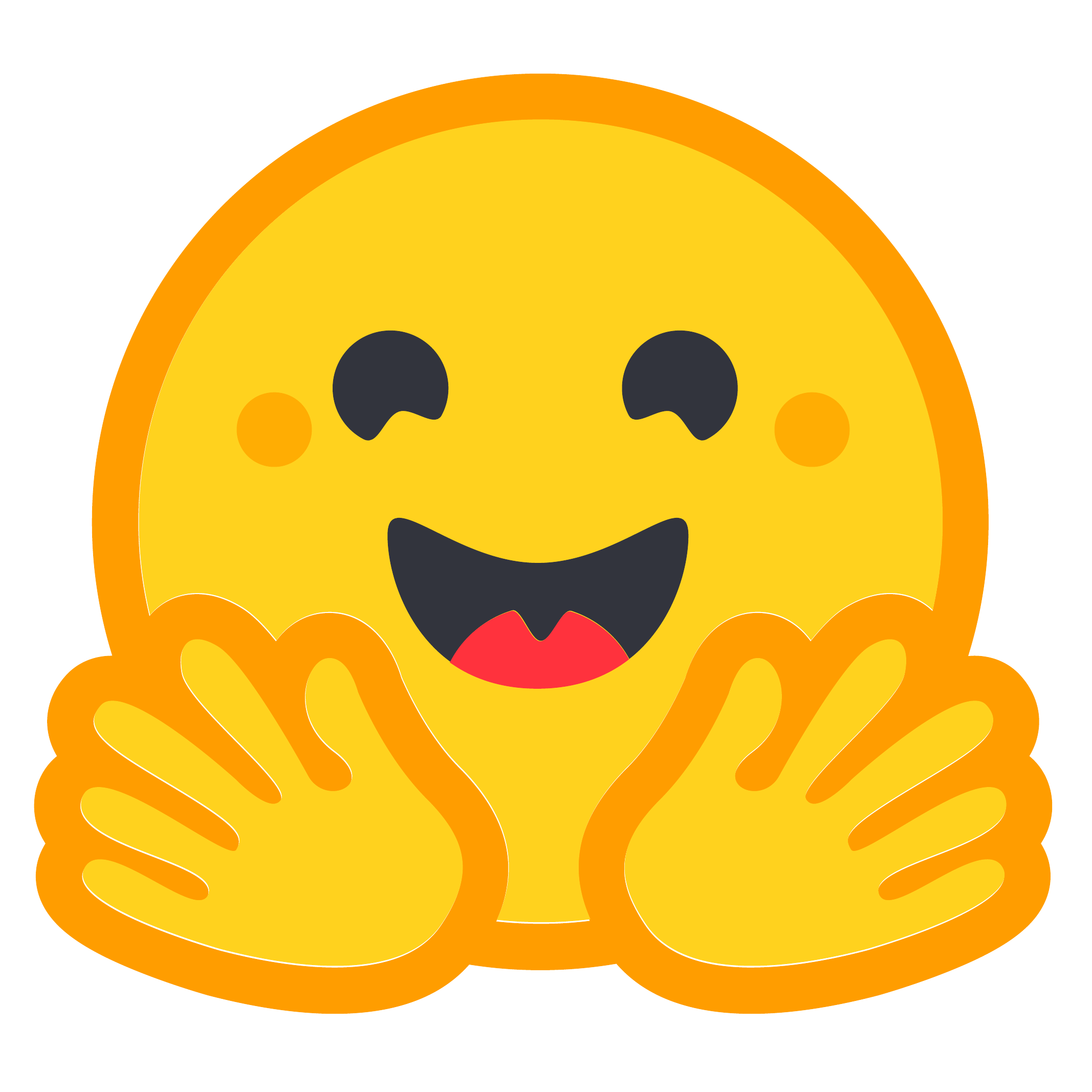}}\xspace}
\newcommand{\coreContrib}{\raisebox{.33em}{\hspace{.05em}\includegraphics[height=.5em]{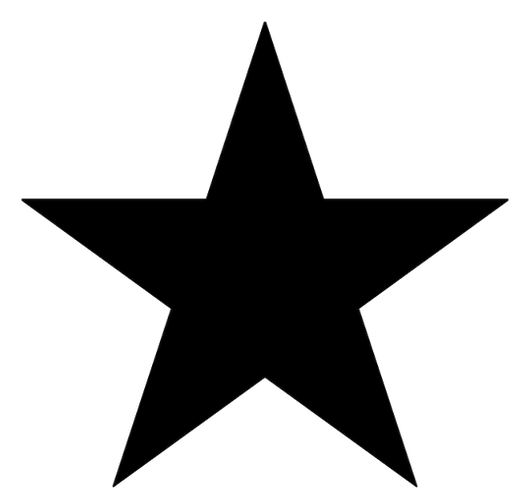}}\xspace}

\newcommand{\github}{\raisebox{-1.5pt}{\includegraphics[height=1.05em]{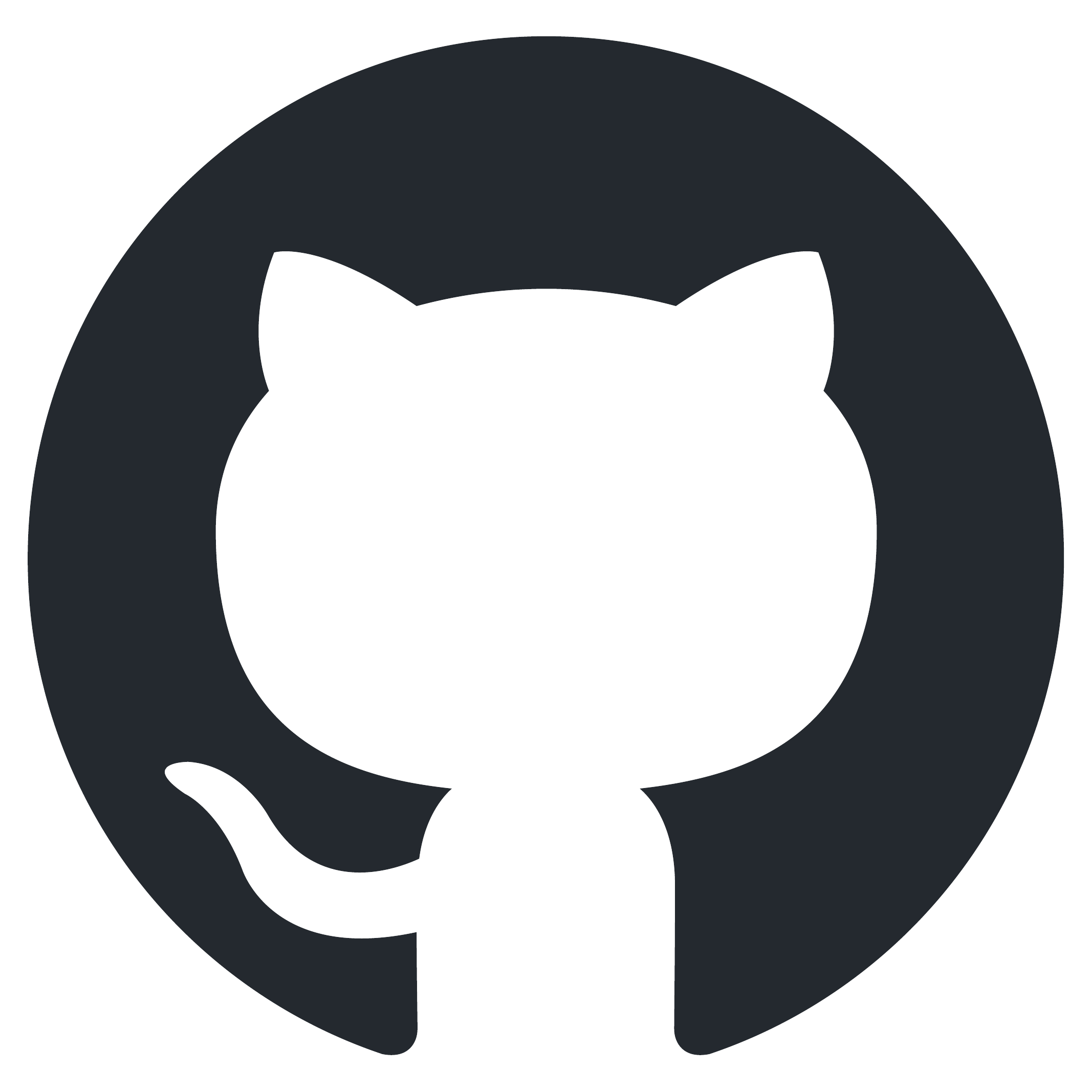}}\xspace}

\newcommand{\spaces}{\raisebox{-1.5pt}{\includegraphics[height=1.05em]{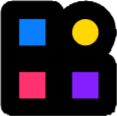}}\xspace}

\newcommand{\chrome}{\raisebox{-1.5pt}{\includegraphics[height=1.05em]{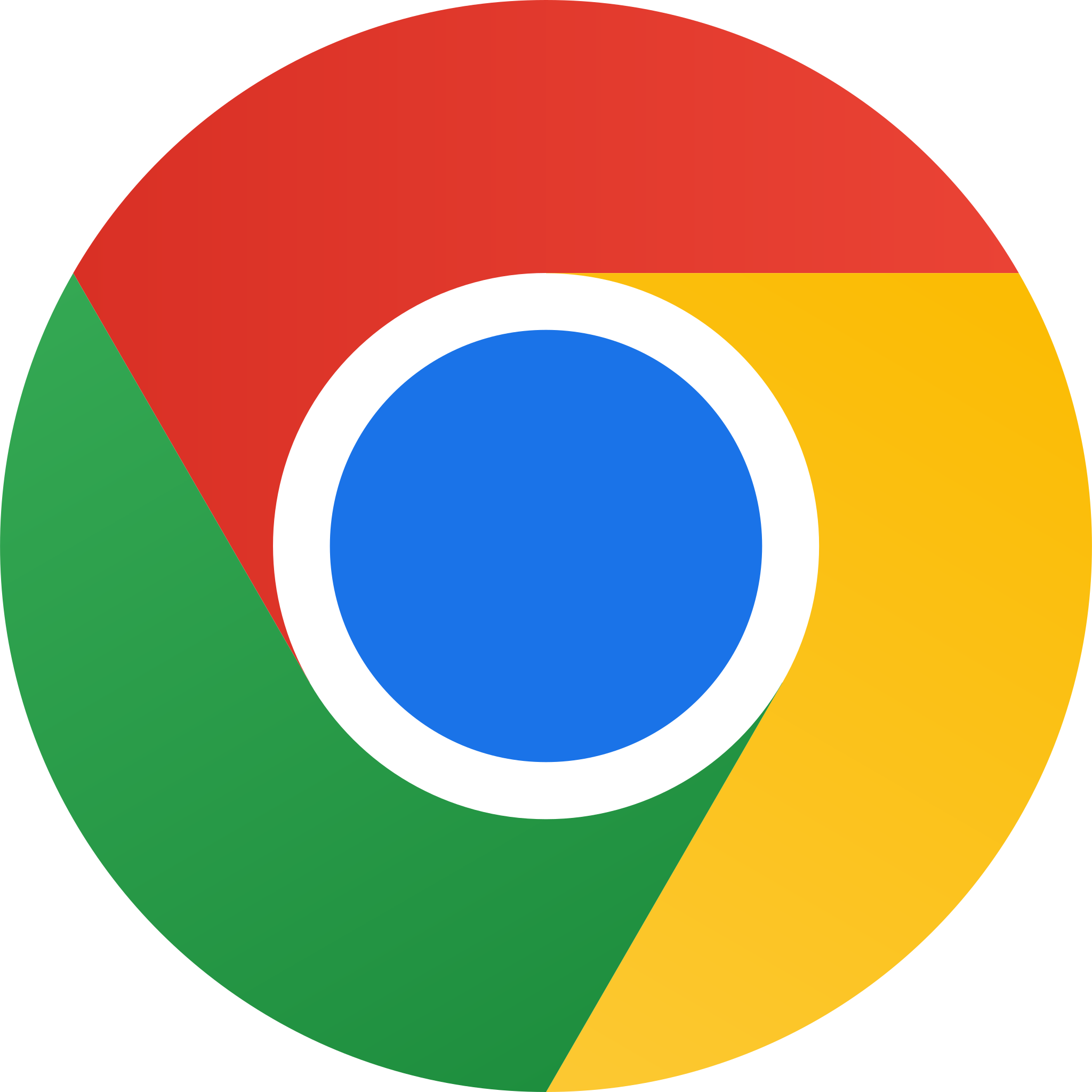}}\xspace}

\newcommand{\huggingsnap}{\raisebox{-1.5pt}{\includegraphics[height=1.05em]{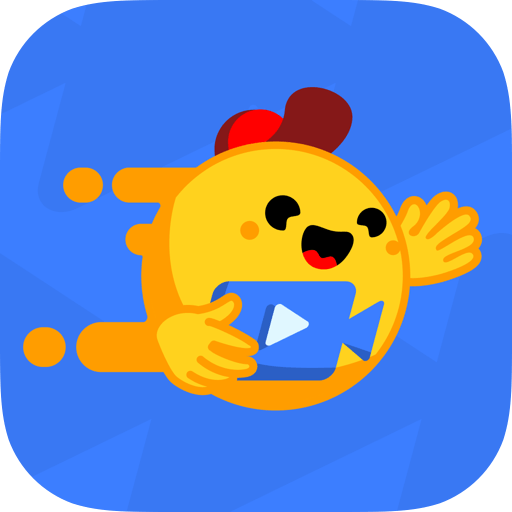}}\xspace}

\newcommand{\stanford}{\raisebox{.28em}{\hspace{.05em}\includegraphics[height=.75em]{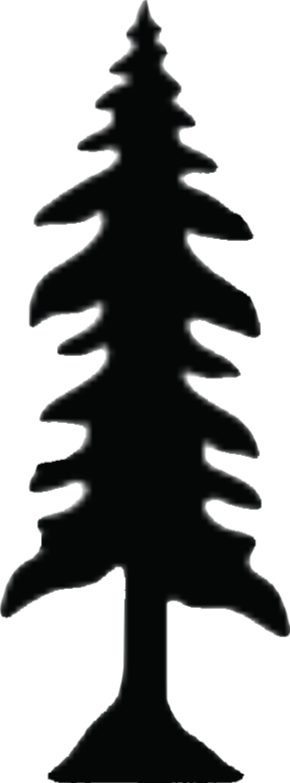}}\xspace}
\newcommand{\hf}{\raisebox{.28em}{\hspace{.05em}\includegraphics[height=.65em]{logos/hf.pdf}}\xspace}

\authorOne[]{Andrés Marafioti\hf\coreContrib}
\authorOne[]{Orr Zohar\stanford\coreContrib}
\authorOne[]{Miquel Farré\hf\coreContrib}

\authorTwo[]{Merve Noyan\hf}
\authorTwo[]{Elie Bakouch\hf}
\authorTwo[]{Pedro Cuenca\hf}
\authorTwo[]{Cyril Zakka\hf}
\authorTwo[]{Loubna Ben Allal\hf}
\authorTwo[]{Anton Lozhkov\hf}
\authorTwo[]{Nouamane Tazi\hf}
\authorTwo[]{Vaibhav Srivastav\hf}
\authorTwo[]{Joshua Lochner\hf}
\authorTwo[]{Hugo Larcher\hf}
\authorTwo[]{Mathieu Morlon\hf}
\authorTwo[]{Lewis Tunstall\hf}
\authorTwo[]{Leandro von Werra\hf}
\authorTwo[]{Thomas Wolf\hf}

\contribution[]{\hf Hugging Face, \stanford Stanford University \quad \coreContrib{ Equal Contribution}}

\setmainfigure{

\begin{figure}[!h]
    \centering
    \vspace{-0.1in}
    \includegraphics[width=0.95\linewidth]{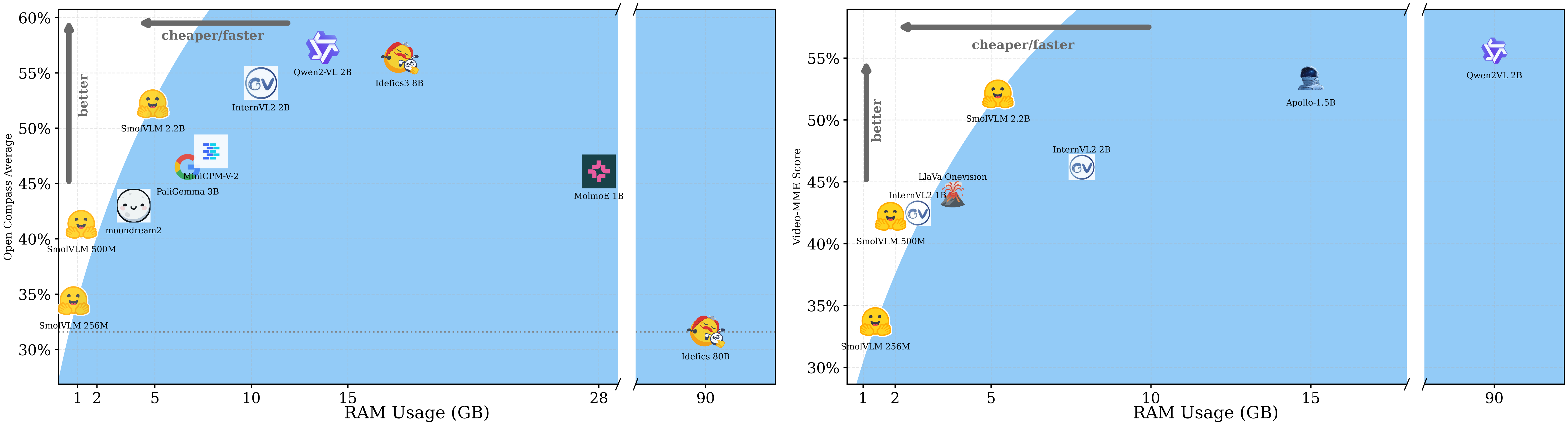}\vspace{-0.1in}
    \caption{\textbf{Smol yet Mighty:} comparison of SmolVLM with other state-of-the-art small VLM models. Image results are sourced from the OpenCompass OpenVLM leaderboard~\citep{VLMEvalKit}.\label{fig:smolvlm_vision_comp}}
\end{figure}

}

\abstract{
Large Vision-Language Models (VLMs) deliver exceptional performance but require significant computational resources, limiting their deployment on mobile and edge devices. Smaller VLMs typically mirror design choices of larger models, such as extensive image tokenization, leading to inefficient GPU memory usage and constrained practicality for on-device applications.

\medskip

We introduce \textbf{SmolVLM}, a series of compact multimodal models specifically engineered for resource-efficient inference. We systematically explore architectural configurations, tokenization strategies, and data curation optimized for low computational overhead. Through this, we identify key design choices that yield substantial performance gains on image and video tasks with minimal memory footprints.

\medskip

Our smallest model, SmolVLM-256M, uses less than 1GB GPU memory during inference and outperforms the 300-times larger Idefics-80B model, despite an 18-month development gap. Our largest model, at 2.2B parameters, rivals state-of-the-art VLMs consuming twice the GPU memory. SmolVLM models extend beyond static images, demonstrating robust video comprehension capabilities.

\medskip

Our results emphasize that strategic architectural optimizations, aggressive yet efficient tokenization, and carefully curated training data significantly enhance multimodal performance, facilitating practical, energy-efficient deployments at significantly smaller scales.
}

\setmaintable{
\begin{table}[!h]
    \centering
    \begin{tabular}{l l @{\hspace{1.5
    cm}} l l}
        \github~{\sans{\textbf{Code}}} & \href{https://github.com/huggingface/smollm}{\texttt{gitub/huggingface}} &
        \huggingface~{\sans \textbf{Blog}} & \href{https://huggingface.co/blog/smolvlm2}{\texttt{blog/smolvlm2}} \\

        \huggingface~{\sans \textbf{Weights}} & \href{https://huggingface.co/HuggingFaceTB}{\texttt{community/smol-research}}  &
        \chrome~{\sans \textbf{VLM Browser}} & \href{https://huggingface.co/spaces/HuggingFaceTB/SmolVLM-500M-Instruct-WebGPU}{\texttt{spaces/smolvlm-webgpu}} \\

        \spaces~{\sans \textbf{Demo}} & \href{https://huggingface.co/spaces/HuggingFaceTB/SmolVLM2}{\texttt{spaces/smolvlm2}}  &
        \huggingsnap~{\sans \textbf{HuggingSnap}} & \href{https://apps.apple.com/us/app/huggingsnap/id6742157364}{\texttt{apple/huggingsnap}} \\
    \end{tabular}
\end{table}
}

\begin{document}

\maketitle

\input{sections/intro}

\input{sections/arch}

\input{sections/sft}

\input{sections/res}

\input{sections/related}

\section{Conclusion}

We introduced \textbf{SmolVLM}, a family of memory-efficient Vision-Language Models ranging from 256M to 2.2B parameters. Remarkably, even our smallest variant requires less than 1GB of GPU memory yet surpasses state-of-the-art 80B-parameter models from just 18 months ago~\citep{OBELICS}. Our findings emphasize a critical insight: scaling down large VLM architectures optimized under resource-rich conditions results in disproportionately high memory demands during inference with little advantage over specialized architectures. By contrast, SmolVLM's design philosophy explicitly prioritizes compact architectural innovations, aggressive but careful tokenization methods, and efficient training strategies, enabling powerful multimodal capabilities at a fraction of the computational cost. 

All model weights, training datasets, and training code are publicly released to encourage reproducibility, transparency, and continued innovation. We hope SmolVLM will inspire the next generation of lightweight, efficient VLMs, unlocking new possibilities for real-time multimodal inference with minimal power consumption.

\newpage

\bibliographystyle{hfstyle/plainnat}
\bibliography{references}

\end{document}

%% file: preamble.tex
\definecolor{lightyellow}{rgb}{1, 0.95, 0.85}
\definecolor{graphicbackground}{rgb}{0.9765,0.9451,0.9059}
\definecolor{codebackground}{rgb}{0.8314,0.949,0.9882}

\newcommand{\finding}[2]{%
  \begin{tcolorbox}[colback=lightyellow, colframe=black, arc=4pt, boxsep=1pt]
    \paragraph{\textbf{\textit{Finding} #1.}} #2
  \end{tcolorbox}%
}

%% file: sections/intro.tex
\begin{figure}[t]
    \centering
    \includegraphics[width=0.89\linewidth]{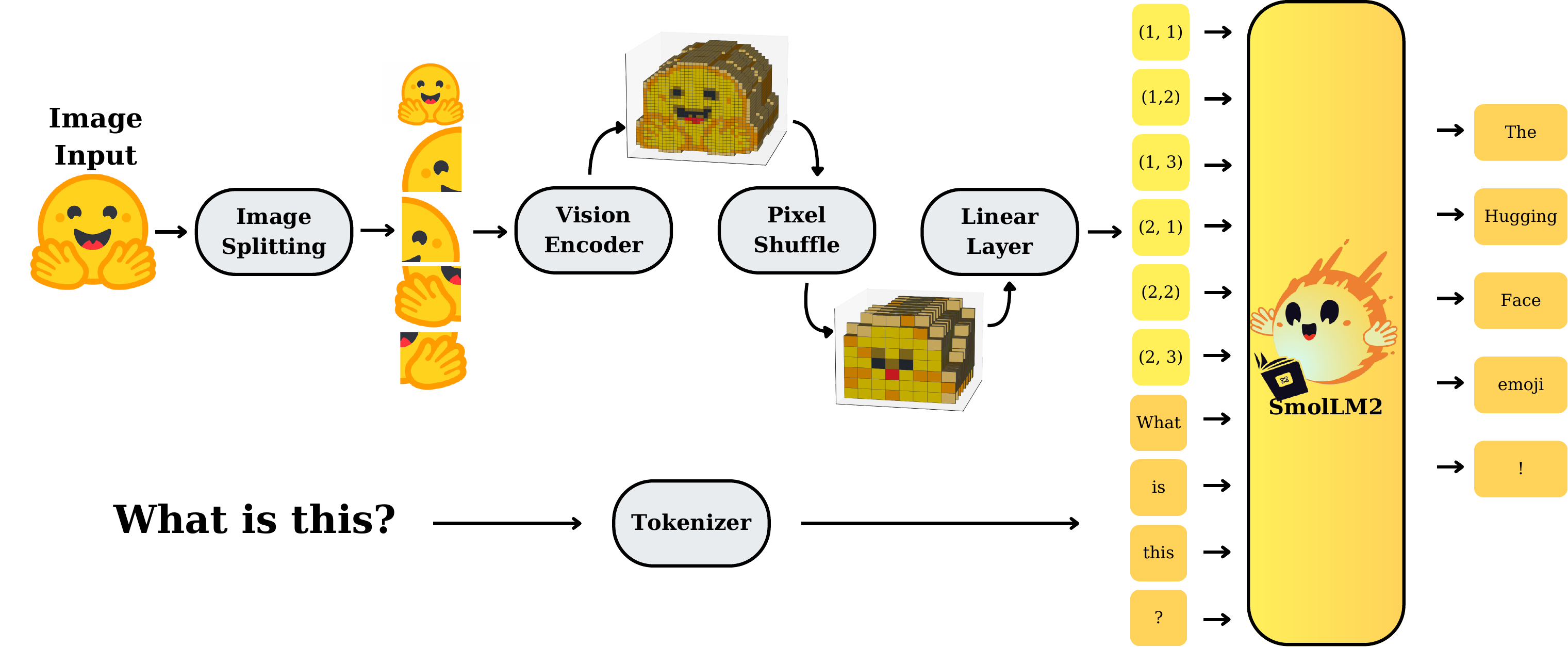}\vspace{-0.09in}
    \caption{\textbf{SmolVLM Architecture}. Images are split into subimages, frames are sampled from videos, and then encoded into visual features. These features are first rearranged via a pixel-shuffle operation, then mapped into the LLM input space as visual tokens using an MLP projection. Visual tokens are then concatenated/interleaved with text embeddings (orange/red). This combined sequence is passed to the LLM for text output.}\vspace{-0.05in}
    \label{fig:smolvlm-arch}
\end{figure}

\section{Introduction}

Vision-Language Models (VLMs) have rapidly advanced in capability and adoption~\citep{GPT4,bai2023qwenvlversatilevisionlanguagemodel,paligemma2024,chen2023internvl,apple-mm1}, driving breakthroughs in cross-modal reasoning~\citep{LLAVA-NeXT,LLaVA} and document understanding~\citep{appalaraju2021docformer,faysse2024colpali,livathinos2025docling,nassar2025smoldoclingultracompactvisionlanguagemodel}. However, these improvements typically entail large parameter counts and high computational demands.

Since early large-scale VLMs like Flamingo~\citep{Flamingo} and Idefics~\citep{OBELICS} demonstrated capabilities with $80$B parameters, new models have slowly appeared at smaller sizes. However, these models often retain high memory demands due to architectural decisions made for their larger counterparts. For instance, Qwen2-VL~\citep{wang2024qwen2} and InternVL 2.5~\citep{chen2024internvl25} offer smaller variants ($1$B-$2$B), but retain significant computational overhead. Conversely, models from Meta~\citep{Llama3} and Google (Gemma 3) reserve vision capabilities for large-scale models. Even PaliGemma~\citep{paligemma2024}, initially efficiency-focused, scaled up significantly in its second release~\citep{paligemma2}. In contrast, Moondream~\citep{moondream} keeps focusing on improving performance while maintaining efficiency, and H$2$OVL-Mississippi~\citep{galib2024h2ovlmississippivisionlanguagemodels} explicitly targets on-device deployment. Efficient processing is particularly critical for video understanding tasks, exemplified by Apollo \citep{zohar2024apollo}, where memory management is essential. Furthermore, reasoning LLMs generate more tokens during inference, compounding computational costs \citep{DeepSeek-R1, OpenAI-o1}. Therefore, efficiency per token becomes vital to ensure models remain practical for real-world use.
\textit{Our contributions are:}

\ifdefined\isarxiv
\begin{itemize}

\item \textbf{Compact yet Powerful Models}: We introduce SmolVLM, a family of powerful small-scale multimodal models, demonstrating that careful architectural design can substantially reduce resource requirements without sacrificing capability.
\item \textbf{Efficient GPU Memory Usage}: Our smallest model runs inference using less than 1GB GPU RAM, significantly lowering the barrier to on-device deployment.
\item \textbf{Systematic Architectural Exploration}: We comprehensively investigate the impact of architectural choices, including encoder-LM parameter balance, tokenization methods, positional encoding, and training data composition, identifying critical factors that maximize performance in compact VLMs.
\item \textbf{Robust Video Understanding on Edge Devices}: We demonstrate that SmolVLM models generalize effectively to video tasks, achieving competitive scores on challenging benchmarks like Video-MME, highlighting their suitability for diverse multimodal scenarios and real-time, on-device applications.
\item \textbf{Fully Open-source Resources}: To promote reproducibility and facilitate further research, we release all model weights, training datasets, and code, including a mobile application showcasing inference on a smartphone.
\end{itemize}

\else

\begin{itemize}

\item \textbf{Compact yet Powerful Models}: We introduce SmolVLM, a family of powerful small-scale multimodal models, demonstrating that careful architectural design can substantially reduce resource requirements without sacrificing capability.

\item \textbf{Efficient GPU Memory Usage}: Our smallest model runs inference using less than 1GB GPU RAM, significantly lowering the barrier to on-device deployment.

\item \textbf{Systematic Architectural Exploration}: We comprehensively investigate the impact of architectural choices, including encoder-LM parameter balance, tokenization methods, positional encoding, and training data composition, identifying critical factors that maximize performance in compact VLMs.

\item \textbf{Robust Video Understanding on Edge Devices}: We demonstrate that SmolVLM models generalize effectively to video tasks, achieving competitive scores on challenging benchmarks like Video-MME, highlighting their suitability for diverse multimodal scenarios and real-time, on-device applications.

\item \textbf{Fully Open-source Resources}: To promote reproducibility and facilitate further research, we release all model weights, datasets, code, and a mobile application showcasing inference on a smartphone.
\end{itemize}

\fi

%% file: sections/arch.tex
\section{Smoller Model Architecture}
\label{sec:smoller_model_arch}
We systematically explore design choices for small multimodal models based on the architecture in Figure~\ref{fig:smolvlm-arch}, where encoded images are pooled and projected into a SmolLM2 backbone. We first analyze optimal compute allocation, showing smaller vision encoders complement compact LMs (\S\ref{subsec:assign_compute}). Extending context length enables higher image resolutions at minimal overhead (\S\ref{subsec:image_encoding_exp}), and pixel shuffling reduces visual tokens further. Finally, we efficiently handle high-resolution images and videos via document-specific image splitting and targeted token compression (\S\ref{subsec:balance_images_videos}). Together, these approaches yield a unified, performant, and cost-effective recipe for tiny LMMs.

\begin{figure}[t]
    \centering
    \includegraphics[width=1\textwidth]{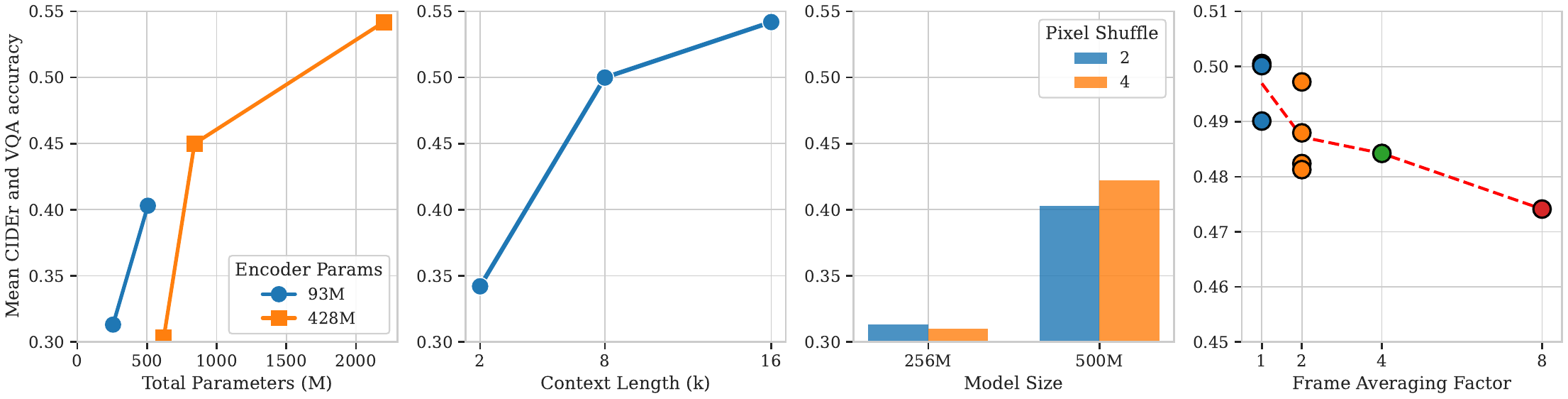}
\caption{\textbf{Performance analysis of SmolVLM configurations.} 
\emph{(Left)} Impact of vision encoder and language model sizes. Smaller language models ($135$M) benefit less from larger vision encoders (SigLIP-SO-$400$M, $428$M) compared to SigLIP-B/$16$ ($93$M), while larger language models gain more from powerful encoders. 
\emph{(Middle-left)} Performance significantly improves with increased context lengths ($2$k to $16$k tokens). 
\emph{(Middle-right)} Optimal pixel shuffle factor (PS=2 vs.\ PS=4) varies by model size. 
\emph{(Right)} Frame averaging reduces video performance, with a rapid decline as more frames are averaged. Metrics average CIDEr (captioning) and accuracy (visual question answering).}

\label{fig:smolvlm_analysis}
\end{figure}

\subsection{How to assign compute between vision and language towers?}
\label{subsec:assign_compute}

VLMs utilize vision encoders (see Figure~\ref{fig:smolvlm-arch}) to generate `vision tokens' that are then fed into an LM.
We investigate optimal capacity allocation between vision encoders and language models (LMs) in compact VLMs. 
Specifically, we pair three SmolLM2 variants ($135$M, $360$M, and $1.7$B parameters) with two SigLIP encoders: a compact $93$M SigLIP-B/$16$ and a larger $428$M SigLIP-SO$400$M. Typically, larger VLMs disproportionately allocate parameters to the LM; 
however, as the LM is scaled down, this is no longer the case. 

Figure~\ref{fig:smolvlm_analysis} (left) confirms that performance declines significantly when using a large encoder with the smallest LM ($135$M), highlighting an inefficient encoder-LM balance. At an intermediate LM scale ($360$M), the larger encoder improves performance by $11.6$\%, yet this comes with a substantial $66$\% increase in parameters, making the compact encoder preferable. Only at the largest LM scale ($1.7$B), the larger encoder represents just a $10$\% parameter increase.

\finding{1}{Compact multimodal models benefit from a balanced encoder-LM parameter allocation, making smaller vision encoders preferable for efficiency.}

\subsection{How can we efficiently pass the images to the Language Model?}

\label{subsec:image_encoding_exp}

Following \citet{Idefics2}, we adopt a self-attention architecture in which visual tokens from the vision encoder are concatenated with textual tokens and jointly processed by a language model (e.g., FROMAGe~\citep{FROMAGe}, BLIP-$2$~\citep{BLIP-2}). This design requires significantly more context than the $2$k-token limit used in SmolLM$2$, as a single $512\times512$ image encoded with SigLIP-B/$16$ requires $1024$ tokens. To address this, we extended the context capacity by increasing the RoPE base from $10$k to $273$k, following \citet{liu2024scalinglawsropebasedextrapolation}, and fine-tuned the model on a mix of long-context data (Dolma books \citep{soldaini2024dolma}, The Stack \citep{kocetkov2022stack}) and short-context sources (FineWeb-Edu \citep{penedo2024refinedweb}, DCLM \citep{li2024datacomp}, and math from SmolLM2).

While fine-tuning was stable at $16$k tokens for the $1.7$B LM, smaller models (135M, $360$M) struggled beyond $8$k. Experiments with our $2.2$B SmolVLM confirmed consistent performance gains up to $16$k tokens (Figure~\ref{fig:smolvlm_analysis}, middle). Accordingly, we adopt a $16$k-token context for SmolVLM and an $8$k-token limit for smaller variants.

\finding{2}{Compact VLMs significantly benefit from extended context lengths.}

Extending the context window alone is not sufficient. Recent VLMs (e.g., MM1~\citep{apple-mm1}, MiniCPM-V~\citep{minicmpv2024}, InternVL~\citep{chen2023internvl}) combine the self-attention architecture with token compression techniques~\citep{zohar2024apollo, laurençon2024buildingbetterunderstandingvisionlanguage} to fit longer sequences efficiently and reduce computational overhead.

\begin{figure}[t]
    \centering
    \includegraphics[width=\linewidth]{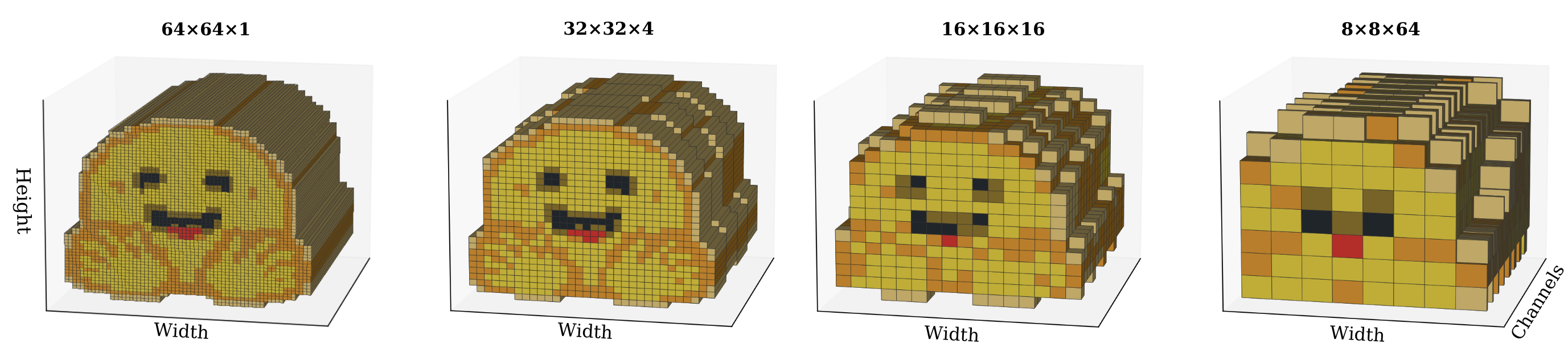}
        \caption{\textbf{Pixel shuffle.} Rearranges encoded images, trading spatial resolution for increased channel depth. This reduces visual token count while preserving information density.}
        \label{fig:pixel_shuffle}
\end{figure}

One particularly effective compression method is \textit{pixel shuffle} (space-to-depth), initially proposed for super-resolution tasks~\citep{shi2016realtimesingleimagevideo} and recently adopted by Idefics3. Pixel shuffle rearranges spatial features into additional channels, reducing spatial resolution but increasing representational density (Figure~\ref{fig:pixel_shuffle}). This reduces the total number of visual tokens by a factor of $r^2$, where $r$ is the shuffle ratio. However, higher ratios collapse larger spatial regions into single tokens, impairing tasks requiring precise localization, such as OCR. Models like InternVL and Idefics$3$ use $r=2$ to balance compression and spatial fidelity. In contrast, our experiments (Figure~\ref{fig:smolvlm_analysis}, right) show that smaller VLMs benefit from more aggressive compression ($r=4$) as the reduced token count eases attention overhead and improves long-context modeling.

\finding{3}{Small VLMs benefit from more aggressive visual token compression.}

\subsection{How can we efficiently encode images and videos?}
\label{subsec:balance_images_videos}

Balancing token allocation between images and videos is crucial for efficient multimodal modeling: images benefit from higher resolution and more tokens to retain fidelity, whereas videos typically require fewer tokens per frame to handle longer sequences efficiently.

To achieve this, we successfully adopted an image-splitting strategy inspired by UReader~\citep{UReader} and SPHINX~\citep{SPHINX}, where high-resolution images are divided into multiple sub-images along with a downsized version of the original. This approach proved effective in maintaining image quality without excessive computational overhead. For videos, however, we found that strategies such as frame averaging, inspired by \cite{liu2024nvila}, negatively impacted performance. As shown in Figure~\ref{fig:smolvlm_analysis} (right), combining multiple frames significantly degraded OpenCompass-Video results, particularly at higher averaging factors ($2$, $4$, $8$). Consequently, frame averaging was excluded from SmolVLM's final design, and video frames were instead rescaled to the resolution of the image encoder.

\finding{4}{For small models, image splitting enhances performance for vision tasks, whereas video frame averaging does not.}

%% file: sections/sft.tex
\section{Smol Instruction Tuning}
\label{sec:tokenization_smoler}

Smol instruction tuning requires careful vision (\S\ref{subsec:learned_vs_str}) and text tokenization (\S\ref{subsec:structured_text}), alongside unified methods for multimodal modeling under tight compute constraints. Learned positional tokens and structured prompts stabilize training and improve OCR, but data composition remains crucial: reusing LLM instruction datasets negatively impacts small VLMs (\S\ref{subsec:text_data_reuse}), excessive Chain-of-Thought data overwhelms limited capacity (\S\ref{subsec:cot_integration}), and moderate video sequence lengths balance efficiency and performance (\S\ref{subsec:video_sequence_length}). Collectively, these insights highlight targeted strategies essential for effectively scaling multimodal instruction tuning to SmolVLMs.

\subsection{Learned Tokens vs. String}
\label{subsec:learned_vs_str}

\begin{figure*}[t]
    \centering
    \includegraphics[width=\textwidth]{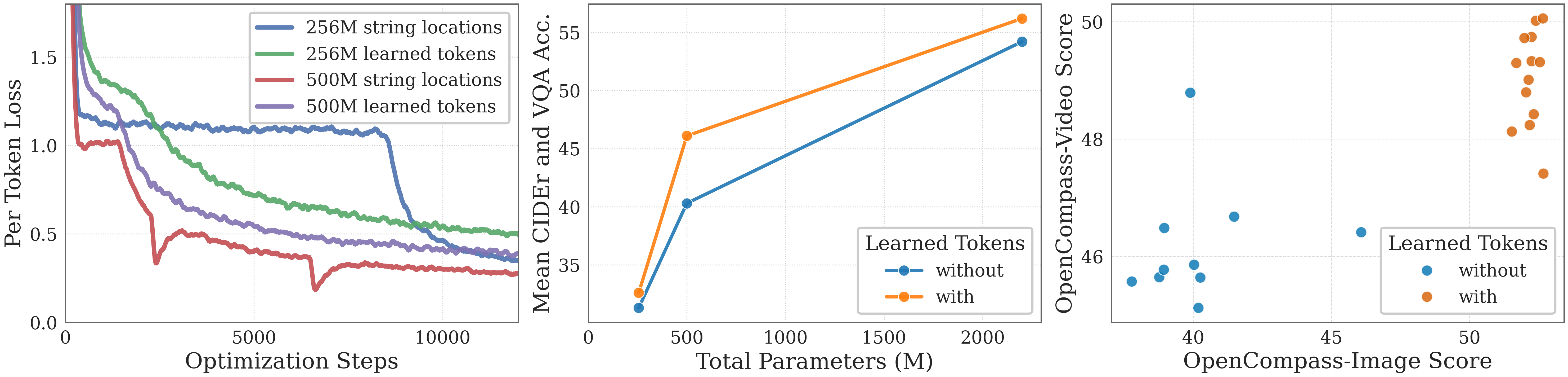}
    \caption{\textbf{Tokenization Strategy Comparisons.}
    \emph{(Left)}~Training loss curves illustrating the ``OCR loss plague'' when using string-based tokens in smaller models. 
    \emph{(Center)}~Aggregated evaluation metrics showing consistently higher scores with learned tokens (orange). 
    \emph{(Right)}~Scatter plot of OpenCompass-Image vs.\ OpenCompass-Video: learned tokens dominate the higher-scoring region, especially in image-intensive tasks.}
    \label{fig:final_token_comparison}
\end{figure*}

A primary design consideration in SmolVLM involves encoding split sub-image positions effectively. Initially, we attempted to use simple string tokens (e.g., \texttt{<row\_1\_col\_2>}), which caused early training plateaus—termed the ``OCR loss plague''—characterized by sudden loss drops without corresponding improvements in OCR performance (Figure~\ref{fig:final_token_comparison}, left and middle).

To address instability during training, we introduced positional tokens, significantly improving training convergence and reducing stalls. Although larger models were relatively robust to using raw string positions, smaller models benefited substantially from positional tokens, achieving notably higher OCR accuracy and improved generalization across tasks. Figure~\ref{fig:final_token_comparison} (center) shows that learned positional tokens consistently outperform naive string positions on multiple image and text benchmarks. Additionally, Figure~\ref{fig:final_token_comparison} (right) illustrates that models leveraging learned tokens consistently score higher in both OpenCompass-Image and OpenCompass-Video evaluations, underscoring the effectiveness of structured positional tokenization in compact multimodal models.

\finding{5}{Learned positional tokens outperform raw text tokens for compact VLMs.}

\subsection{Structured Text Prompts and Media Segmentation}
\label{subsec:structured_text}
\begin{figure*}[t]
    \centering
    \includegraphics[width=\textwidth]{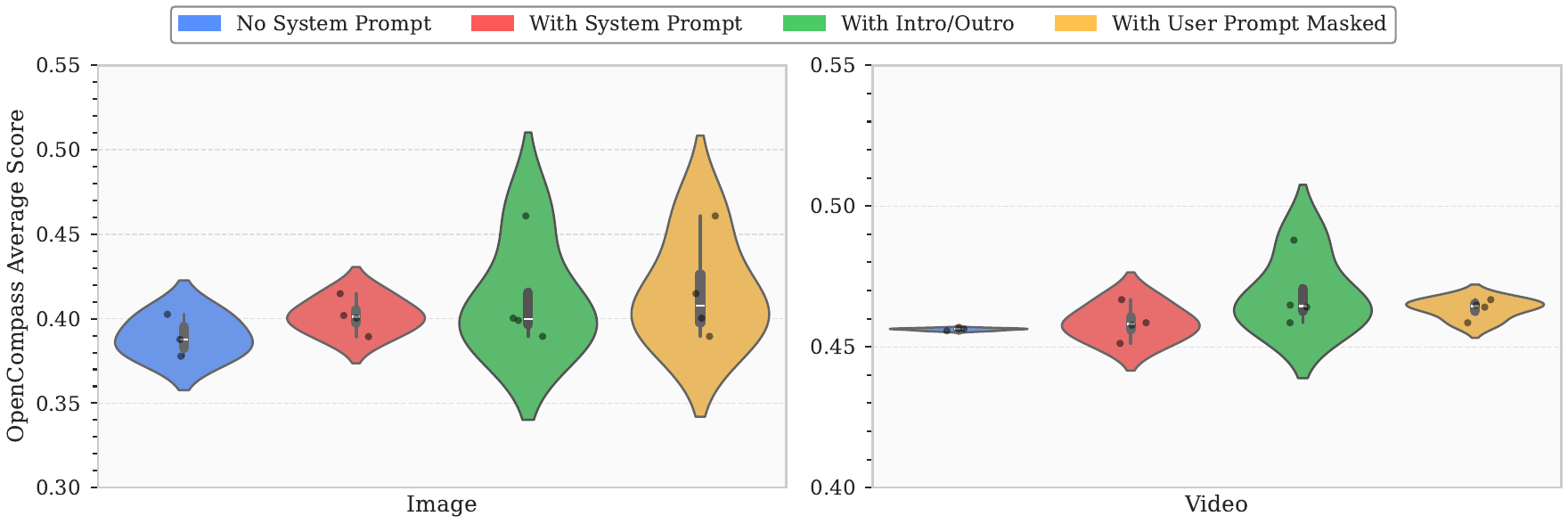}
        \caption{
    \textbf{Cumulative Effect of Training Strategies on SmolVLM Performance.}
    The visualization shows the progression of performance improvements as different tokenization and prompt engineering strategies are applied sequentially to the SmolVLM base model. 
    \emph{(Left)}~Image benchmark results show consistent improvements with each added strategy. 
    \emph{(Right)}~Video benchmark results reveal similar patterns with more pronounced gains.
    }
    \label{fig:violin_comparison}
\end{figure*}

We evaluated how system prompts and explicit media intro/outro prefixes incrementally improve SmolVLM’s performance on image (left) and video (right) benchmarks, as shown in Figure~\ref{fig:violin_comparison}. Each violin plot represents three checkpoints for a given configuration.

\textbf{System Prompts.} 
We prepend concise instructions to clarify task objectives and reduce ambiguity during zero-shot inference. For example, conversational datasets utilize prompts like \textit{``You are a useful conversational assistant,''} whereas vision-focused tasks employ \textit{``You are a visual agent and should provide concise answers.''} The second violin plot in each subplot (Fig.~\ref{fig:violin_comparison}) illustrates clear performance improvements from incorporating these system prompts, particularly evident in image-centric tasks.

\textbf{Media Intro/Outro Tokens.} 
To clearly demarcate visual content, we introduce textual markers around image and video segments (e.g., ``\textit{Here is an image...}'' and ``\textit{Here are $N$ frames sampled from a video...}''). The outro tokens then transition back to textual instructions (e.g., ``\textit{Given this image/video...}''). The third violin indicates that this strategy substantially boosts performance on video tasks—where confusion between multiple frames is more likely—and still yields measurable improvements on image tasks.

\textbf{Masking User Prompts}
\label{subsec:masking_user_prompts}
Drawing on techniques from \citet{allal2025smollm2smolgoesbig}, we explore user-prompt masking during supervised fine-tuning as a way to reduce overfitting. The right violin plot in Figure~\ref{fig:violin_comparison} shows that masking user queries (orange) yields improved performance in both image and video tasks, compared to the unmasked baseline (blue). This effect is significantly pronounced in multimodal QA, where questions are often repetitive and can be trivially memorized by the model. Masking thus forces SmolVLM to rely on task-related content rather than superficial repetition, promoting better generalization.

\finding{6}{System prompts and media intro/outro tokens significantly improve compact VLM performance, particularly for video tasks. During SFT, only train on completions.}

\subsection{Impact of Text Data Reuse from LLM-SFT}
\label{subsec:text_data_reuse}
A seemingly intuitive practice is to reuse text data from the final supervised fine-tuning stages of large language models, anticipating in-distribution prompts and higher-quality linguistic inputs. However, Figure~\ref{fig:training_strategies} (left) shows that incorporating LLM-SFT text data (\emph{SmolTalk}) can degrade performance in smaller multimodal architectures by as much as 3.7\% in video tasks and 6.5\% in image tasks. We attribute this negative transfer to reduced data diversity, which outweighs any benefits of reusing text. In keeping with \citet{zohar2024apollo}, we therefore maintain a strict 14\% text proportion in our training mix. These findings highlight the importance of a carefully balanced data pipeline, rather than direct adoption of large-scale SFT text for small-scale multimodal models.

\finding{7}{Adding text from SFT blend proved worse than new text SFT data.}

\subsection{Optimizing Chain-of-Thought Integration for Compact Models}
\label{subsec:cot_integration}

Chain-of-Thought (CoT) prompting, which exposes models to explicit reasoning steps during training, generally enhances reasoning capabilities in large models. However, its effect on smaller multimodal architectures remains unclear. To investigate this, we varied the proportion of CoT data integrated into the Mammoth dataset \citep{yue2023mammoth}, covering text, image, and video tasks. Figure~\ref{fig:training_strategies} (middle) shows that incorporating a minimal fraction (0.02–0.05\%) of CoT examples slightly improved performance, but higher proportions markedly degraded results, especially in image tasks. These observations suggest that excessive reasoning-oriented textual data can overwhelm the limited capacity of smaller VLMs, thereby compromising their visual representation capabilities. Consequently, compact models benefit most from very sparse inclusion of CoT data rather than the extensive use typically beneficial in larger-scale architectures.

\finding{8}{Excessive CoT data harms compact model performance.}

\subsection{Impact of Video Sequence Length on Model Performance}
\label{subsec:video_sequence_length}

Increasing video duration during training offers richer temporal context but comes at a greater computational cost. To identify an optimal duration, we trained SmolVLM on average video lengths ranging from 1.5 to 3.5 minutes. Figure~\ref{fig:training_strategies} (right) demonstrates clear performance improvements for both video and image benchmarks as video durations approached approximately 3.5 minutes, likely due to more effective cross-modal feature learning. Extending video duration beyond 3.5 minutes yielded minimal further gains, indicating diminishing returns relative to the added computational expense. Thus, moderately extending video sequences enhances performance significantly in smaller models, whereas overly long sequences do not proportionally justify their computational cost.

\finding{9}{Moderately increasing video duration during training improves both video and image task performance in compact VLMs.}

%% file: sections/res.tex
\begin{figure}[t]
    \centering
    \includegraphics[width=\textwidth]{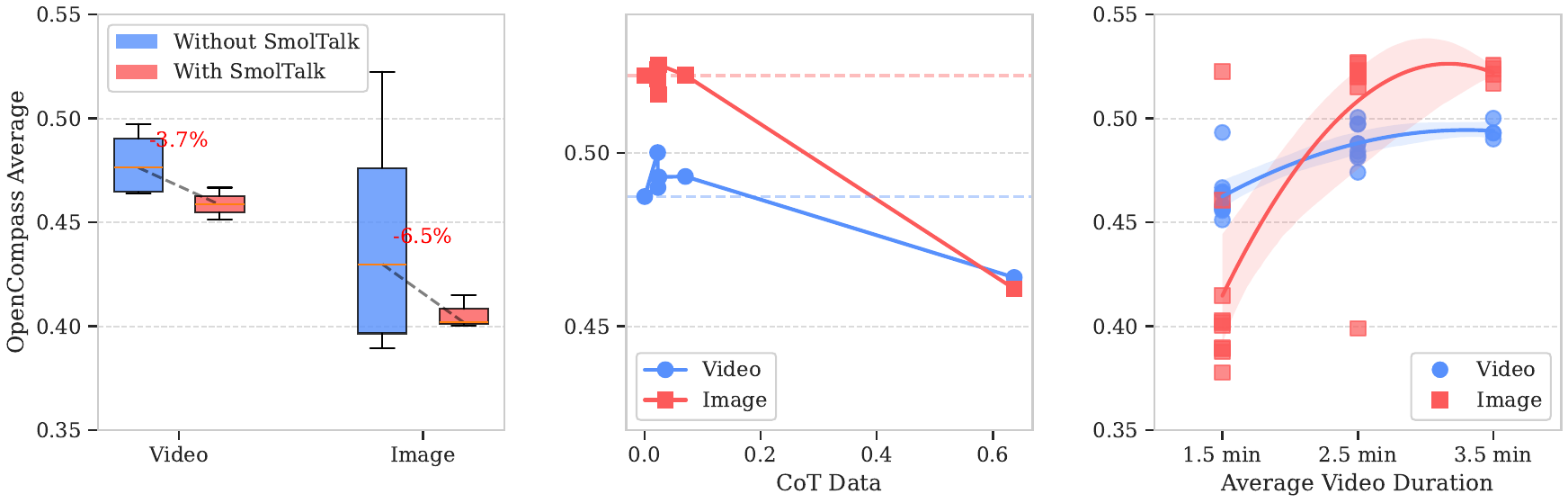}
    \caption{\textbf{Impact of Training Strategies on Smol-Scale Multimodal Models.}
    \emph{(Left)}~Reusing text data from LLM-SFT (\emph{SmolTalk}) reduces both image and video scores in smaller models.
    \emph{(Middle)}~A minimal fraction ($0.02$\%--$0.05$\%) of Chain-of-Thought (CoT) data yields optimal results, while heavier CoT usage degrades performance.
    \emph{(Right)}~Increasing average video duration beyond $3.5$ min leads to diminished returns for both image and video tasks.
    }
    \label{fig:training_strategies}
\end{figure}

\section{Experimental Results}

We construct three variants of SmolVLM, tailored to different computational environments:
\begin{itemize}
    \item \textbf{SmolVLM-$\bm{256}$M}: Our smallest model, combining the $93$M SigLIP-B/$16$ and the SmolLM$2$-$135$M~\citep{allal2025smollm2smolgoesbig}. Operating on $<1$GB GRAM makes it ideal for resource-constrained edge applications.
    \item \textbf{SmolVLM-$\bm{500}$M}: A mid-range model with the same 93M SigLIP-B/$16$ paired with the larger SmolLM$2$-$360$M. Balancing memory efficiency and performance, it is suitable for moderate-resource edge devices.
    \item \textbf{SmolVLM-$\bm{2.2}$B}: The largest variant, with a $400$M SigLIP-SO$400$M and a $1.7$B-parameter SmolLM$2$ backbone. This model maximizes performance while remaining deployable on higher-end edge systems.
\end{itemize}

\subsection{Training Data}

\label{subsec:training_data}

Model training proceeds in two stages: (1) a vision stage, and (2) a video stage. The vision training stage uses a new mixture of the datasets used in \cite{laurençon2024buildingbetterunderstandingvisionlanguage}, to which we added MathWriting \citep{mathwritingdataset}. The mixture was balanced to emphasize visual and structured data interpretation while maintaining the focus on reasoning and problem-solving capabilities. The visual components comprise document understanding, captioning, and visual question answering (including $2$\% dedicated to multi-image reasoning), chart understanding, table understanding, and visual reasoning tasks. To preserve the model's performance in text-based tasks, we retained a modest amount of general knowledge Q\&A and text-based reasoning \& logic problems, which incorporate mathematics and coding challenges.

The video fine-tuning stage maintains $14$\% of text data and $33$\% of video to achieve optimal performance, following the learnings of \cite{zohar2024apollo}. For video, we sample visual description and captioning from LLaVA-video-$178$k~\citep{llava-video-178k}, Video-STAR~\citep{video-star}, Vript~\citep{Vript}, and ShareGPT4Video~\citep{ShareGPT4V}, temporal understanding from Vista-$400$k~\citep{vista-400k}, and narrative comprehension from MovieChat~\citep{moviechat} and FineVideo~\citep{FineVideo}. Multi-image data was sampled from M$4$-Instruct~\citep{LLAVA-NeXT} and Mammoth~\citep{mammoth}. The text samples were sourced from~\citep{xu2024magpie}. 

\begin{figure}[t]
    \centering
    \includegraphics[width=0.45\textwidth]{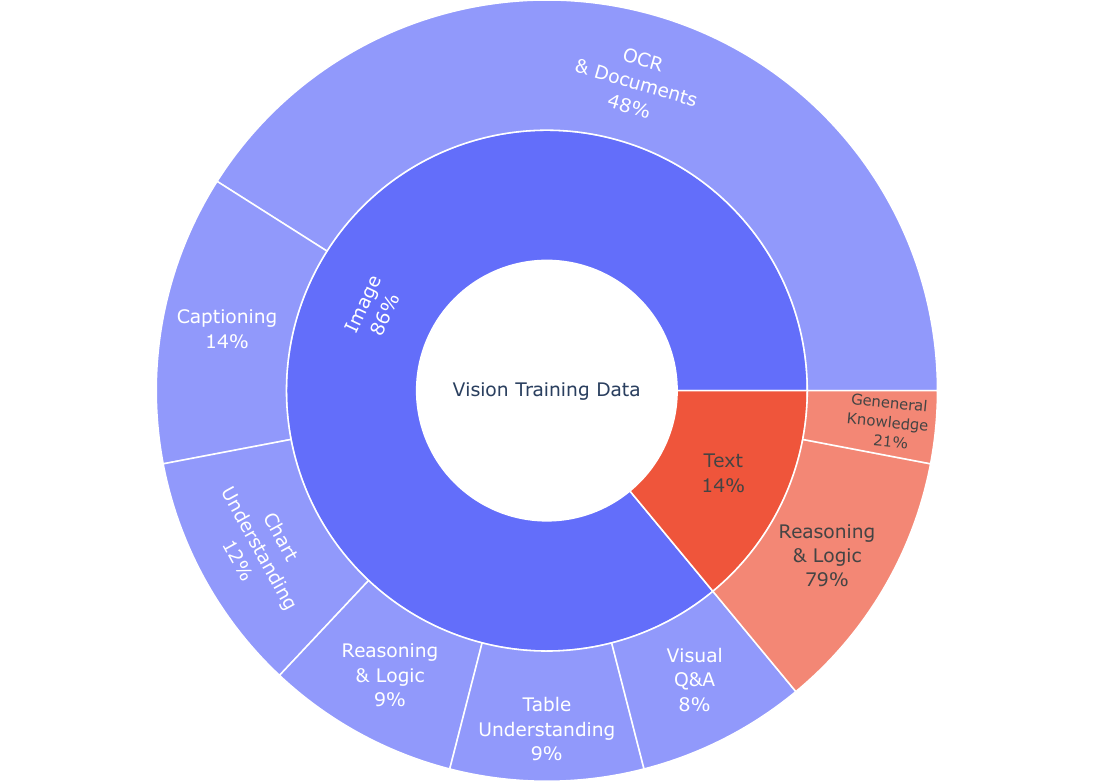}
    \hspace{0.05in}
    \includegraphics[width=0.45\textwidth]{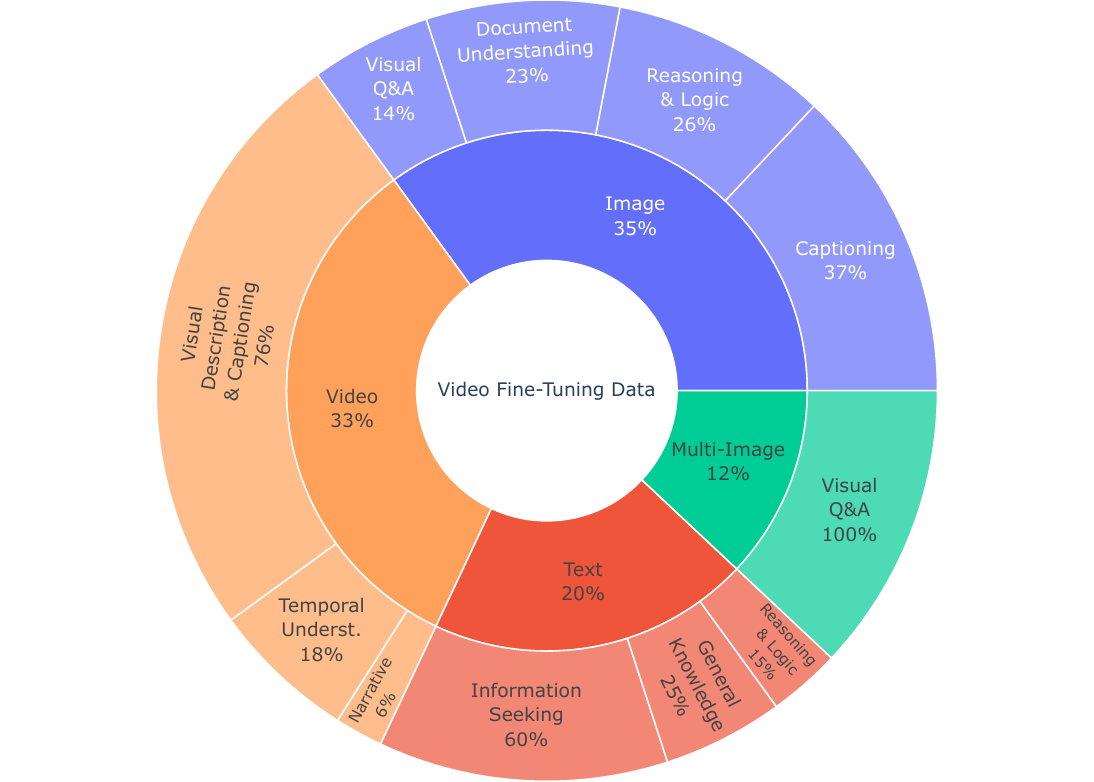}
    \caption{\textbf{Data Details.} Training dataset details for Vision  \emph{(Left)} and video \emph{(Right)}, broken down by modality and sub-categories.}
    \label{fig:data_slices}
\end{figure}

For a more granular description, Figure~\ref{fig:data_slices} provides a detailed overview of the training data distribution used in both our vision and video fine-tuning stages.

\subsection{Evaluation details}

We evaluated SmolVLM using VLMEvalKit \citep{VLMEvalKit} to ensure reproducibility. The full results are available online\footnote{\href{https://huggingface.co/spaces/opencompass/open\_vlm\_leaderboard}{OpenVLM Leaderboard}}. Currently, the OpenVLM Leaderboard covers $239$ different VLMs and $31$ different multi-modal benchmarks. 
Further, we plot the performance against the RAM required to run the evaluations. We argue that model size is usually used as a proxy for the computational cost required to run a model. This is misleading for VLMs because the architecture strongly influences how expensive it is to run the model; in our opinion, RAM usage is a better proxy. For SmolVLM, this resizes the longest edge of images to $1920$ in the $256$M and $500$M models and $1536$ in the $2.2$B.

\subsection{Strong Performance at a Tiny Scale}
\label{subsec:perf_vs_size}

We evaluate SmolVLM's performance relative to model size, comparing three variants ($256$M, $500$M, and $2.2$B) against efficient state-of-the-art open-source models. Table~\ref{tab:smolvlm_vision_video_size_comp} summarizes results across nine demanding vision-language benchmarks and five video benchmarks. We highlight in the table MolmoE $7$B with $1$B activated parameters~\citep{deitke2024molmopixmoopenweights}(MolmoE-A$1$B-$7$B) for vision tasks and InternVL$2$-$2$B~\cite{chen2023internvl} for video tasks. A broader array of competing models are shown in Fig.~\ref{fig:smolvlm_vision_comp}.

\input{tables/llava_onevision_table}

\paragraph{Efficiency and Memory Footprint.}
SmolVLM demonstrates remarkable computational efficiency compared to significantly larger models. Single-image inference requires only $0.8$GB of VRAM for the $256$M variant, $1.2$GB for the $500$M, and $4.9$GB for the $2.2$B—dramatically lower than the $27.7$GB required by MolmoE-A$1$B-$7$B. Even compared to models of similar parameter scales, SmolVLM is notably more efficient: Qwen$2$VL-$2$B requires $13.7$GB VRAM and InternVL2-$2$B requires $10.5$GB VRAM, highlighting that parameter count alone does not dictate compute requirements. At batch size $64$, memory usage 
for SmolVLM remains practical: $15.0$GB ($256$M), $16.0$GB ($500$M), and $49.9$GB ($2.2$B). These results highlight SmolVLM's substantial advantages for deployment in GPU-constrained environments.

\paragraph{Overall Gains from Scaling.}
Increasing SmolVLM's parameter count consistently yields substantial performance improvements across all evaluated benchmarks. The largest model ($2.2$B) achieves the highest overall score at $59.8\%$, followed by the intermediate $500$M variant ($51.0$\%) and the smallest $256$M variant ($44.0\%$). Notably, even the smallest SmolVLM-$256$M significantly surpasses the much larger Idefics~$80$B model (see Fig.~\ref{fig:smolvlm_vision_comp}) on nearly all benchmarks, emphasizing effective vision capabilities at modest scales. The few exceptions—particularly MMMU ($29.0\%$ vs. $42.3\%$) and AI2D ($46.4\%$ vs. $56.3\%$)—highlight benchmarks where strong linguistic reasoning from a large language backbone remains crucial. Intriguingly, visually oriented tasks such as OCRBench also benefit markedly from scaling language model capacity, with a nearly $10$-point improvement when moving from $256$M ($52.6$\%) to $500$M ($61.0$\%). These results underscore that larger language models provide enhanced context management and improved multimodal reasoning, benefiting both language-intensive and vision-centric tasks.

\paragraph{Comparison with Other Compact VLMs.}
Figure~\ref{fig:smolvlm_vision_comp} situates SmolVLM-$2.2$B among recent small-scale VLMs by comparing OpenCompass benchmark performance against GPU memory consumption per image. SmolVLM-$2.2$B achieves notably strong performance on MathVista ($51.5$) and ScienceQA ($90.0$), while maintaining exceptionally low GPU usage of just $4.9$GB VRAM. In contrast, models requiring significantly more compute, such as Qwen$2$VL-$2$B and InternVL$2$-$2$B, aren't clearly better performers. Specifically, Qwen$2$VL-$2$B slightly surpasses SmolVLM-$2.2$B on AI$2$D ($74.7$ vs. $70.0$) and ChartQA ($73.5$ vs. $68.8$), yet falls short on MathVista ($48.0$ vs. $51.5$) and ScienceQA ($78.7$ vs. $90.0$). Similarly, InternVL2-$2$B achieves higher scores on ScienceQA ($94.1$ vs. $90.0$) and MMStar ($49.8$ vs. $46.0$), but at more than double the VRAM cost.

Further comparisons highlight distinct trade-offs among size, memory footprint, and task-specific performance. MiniCPM-V2 ($2.8$B parameters) underperforms SmolVLM-$2.2$B on most benchmarks. Other models such as Moondream2 and PaliGemma (both around $2–3$B parameters) exhibit significant variance across tasks: Moondream2, for instance, scores well on ChartQA ($72.2$) with just $3.9$GB VRAM but substantially underperforms on MMMU ($29.3$). Conversely, PaliGemma excels at ScienceQA ($94.3$) yet struggles on ChartQA ($33.7$). This variability underscores how specialized training impacts per-task.

\paragraph{Video Benchmarks.}
Table~\ref{tab:smolvlm_vision_video_size_comp} provides comprehensive results across five diverse video benchmarks: Video-MME, MLVU, MVBench, TempCompass, and WorldSense. SmolVLM-$2.2$B notably excels at Video-MME ($52.1$) and WorldSense ($36.2$), outperforming significantly larger models such as Qwen$2$ VL-7B ($32.4$ on WorldSense), showcasing strong capabilities in complex multimodal video comprehension tasks. The SmolVLM-$500$M variant also demonstrates robust performance, achieving competitive scores on TempCompass ($49.0$) and WorldSense ($30.6$), highlighting sophisticated temporal reasoning and real-world visual understanding at a scale ideal for edge-device deployment. Despite their compact parameter counts, SmolVLM variants consistently balance efficient resource use with impressive accuracy, reinforcing their suitability for resource-constrained scenarios.

\begin{figure}[t]
    \centering
    \includegraphics[width=\textwidth, trim=0 2pt 0 2pt, clip]{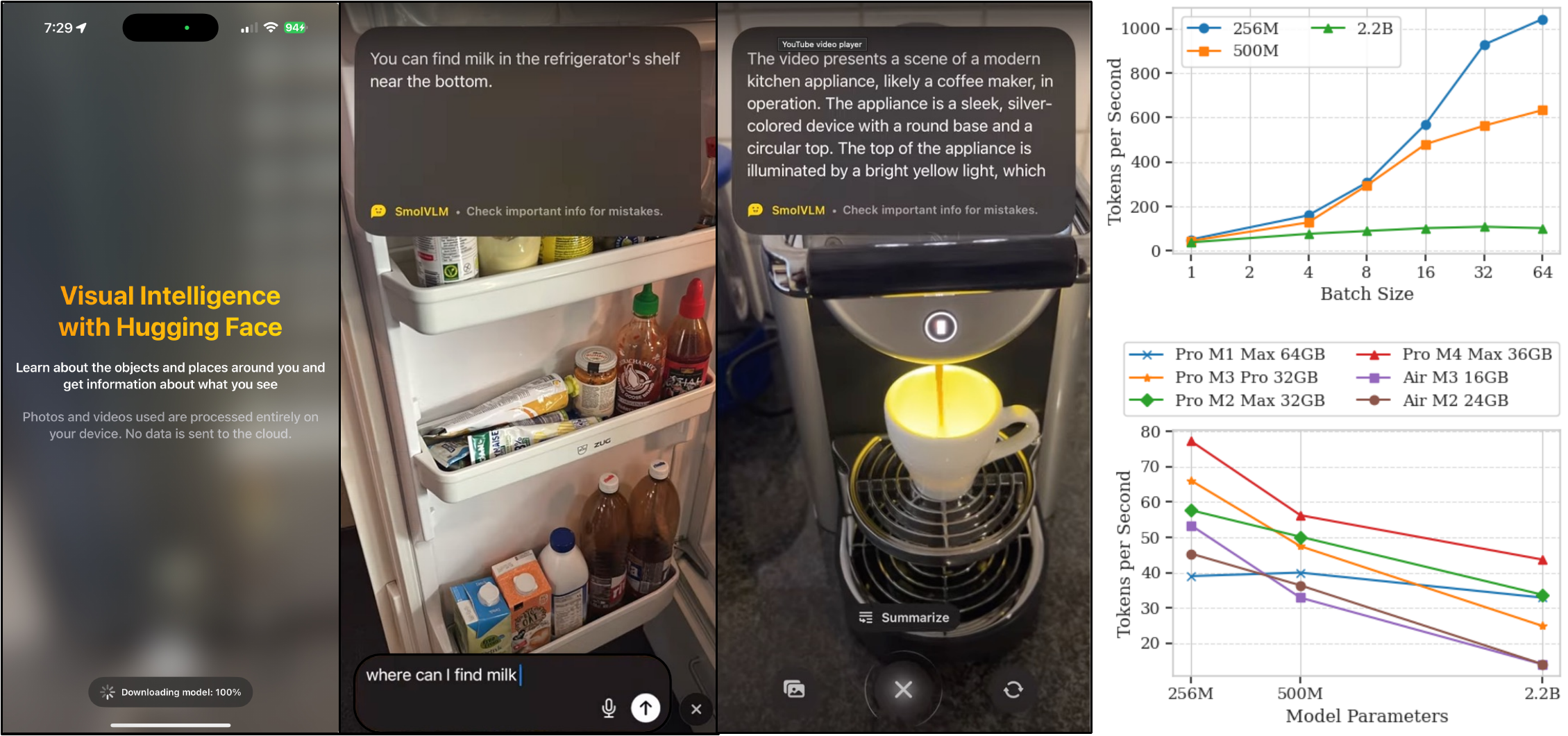}
     \caption{\textbf{SmolVLM on edge device.} 
     \textit{(Left)} Examples of the 
     \href{https://apps.apple.com/us/app/huggingsnap/id6742157364}{\texttt{HuggingSnap}}
     app, where SmolVLM can run locally, on the device, on consumer phones. For example, interactions can be done using a mobile interface to detect objects and answer questions. 
     \textit{(Right)} Throughput in tokens per second on NVIDIA A$100$ GPUs \emph{(top)} and different consumer personal computers \emph{(bottom)} across different batch sizes and model variants.}
       \label{fig:on_device_performance}
\end{figure}

\subsection{On-Device Performance}

To comprehensively assess the deployment practicality of SmolVLM, we benchmarked its throughput across varying batch sizes on two representative hardware platforms: NVIDIA A$100$ and NVIDIA L$4$ GPUs (see Figure~\ref{fig:on_device_performance}). Our evaluations highlight SmolVLM’s suitability for on-device and edge deployment scenarios.

On the A$100$ GPU, the smallest SmolVLM-256M variant achieves impressive throughput, scaling from $0.8$ examples per second at batch size $1$ to $16.3$ examples per second at batch size $64$. The $500$M variant similarly scales from $0.7$ to $9.9$ examples per second, while the largest 2.2B variant demonstrates more modest scaling ($0.6$ to $1.7$ examples per second), indicative of its higher computational demands.

Evaluations on the L$4$ GPU further emphasize SmolVLM’s edge compatibility. Here, the $256$M variant reaches peak throughput at $2.7$ examples per second with batch size $8$, subsequently diminishing due to memory constraints. The $500$M and $2.2$B variants peak at lower batch sizes ($1.4$ and $0.25$ examples per second, respectively), underscoring their efficiency even under more restrictive hardware conditions.

Finally, we accompany the release with several optimized ONNX (Open Neural Network Exchange) exports, facilitating cross-platform compatibility and broadening deployment opportunities across consumer-grade hardware targets. Notably, we demonstrate the ability to efficiently run these models locally within a browser environment via WebGPU, with the $256$M variant achieving up to $80$ decode tokens per second on a $14$-inch MacBook Pro (M$4$ Max).

\subsection{Downstream Applications}

Beyond our own evaluations, SmolVLM has seen adoption in various downstream applications developed by the broader research community, emphasizing its efficiency in real-world, resource-constrained scenarios.

\paragraph{ColSmolVLM: On-Device Multimodal Inference.}
ColSmolVLM utilizes the smaller SmolVLM variants ($256$M and $500$M parameters) designed explicitly for on-device deployment, as detailed in recent work by Hugging Face~\citep{faysse2024colpaliefficientdocumentretrieval}. These compact models enable efficient multimodal inference directly on mobile devices, consumer laptops, and even within browser-based environments, significantly lowering computational demands and operational costs.

\paragraph{Smol Docling: Ultra-Compact Document Processing.}
Smol Docling is an ultra-compact 256M-parameter variant of SmolVLM, optimized explicitly for end-to-end multimodal document conversion tasks~\citep{smoldocling}. By employing specialized representations known as DocTags, Smol Docling efficiently captures content, context, and spatial relationships across diverse document types, including business documents, academic papers, and patents. Its compact architecture maintains competitive performance with considerably larger VLMs, highlighting its suitability for deployment in scenarios with computational constraints.

\paragraph{BioVQA: Biomedical Visual Question Answering.}
BioVQA leverages SmolVLM’s compact and efficient architecture to address visual question answering tasks within the biomedical domain~\citep{lozano2025largescalevisionlanguagedatasetderived}. Small-scale SmolVLM models have demonstrated promising capabilities in interpreting medical images, assisting healthcare professionals by providing accurate answers to clinical questions based on visual data. This capability is particularly valuable in healthcare settings where quick, reliable image interpretation is critical, yet computational resources may be limited.

%% file: tables/llava_onevision_table.tex
\definecolor{light-gray}{gray}{0.6}
\definecolor{front-color}{HTML}{F5FFFA}
\begin{table}[t!]
    \centering
    \setlength{\tabcolsep}{4pt}
    \renewcommand{\arraystretch}{1.1}
    \scriptsize
    \adjustbox{width=\linewidth}{
    \begin{tabular}{p{1.4cm}p{4cm}ccc>{\centering\arraybackslash}p{1.9cm}}
    \toprule
    \addlinespace[2pt]
    \textbf{Capability} & \textbf{Benchmark} & \textbf{SmolVLM 256M} & \textbf{SmolVLM 500M} & \textbf{SmolVLM 2.2B} & \textbf{Efficient OS} \\
    \addlinespace[2pt]
    \midrule
    \multirow{6}{*}{Single-Image} 
     & OCRBench~\citep{liu2024ocrbench} \newline \tiny{\color{light-gray}{Character Recognition}} & \cellcolor{front-color}52.6\% & \cellcolor{front-color}61.0\% & \cellcolor{front-color}72.9\% & 54.7\% \ \tiny{\color{light-gray}{MolmoE-A1B-7B}} \\
     & AI2D~\citep{AI2D} \newline \tiny{\color{light-gray}{Science Diagrams}} & \cellcolor{front-color}46.4\% & \cellcolor{front-color}59.2\% & \cellcolor{front-color}70.0\% & 71.0\%  \ \tiny{\color{light-gray}{MolmoE-A1B-7B}}  \\
     & ChartQA~\citep{ChartQA} \newline \tiny{\color{light-gray}{Chart Understanding}} & \cellcolor{front-color}55.6\% & \cellcolor{front-color}62.8\% & \cellcolor{front-color}68.7\% & 48.0\%  \ \tiny{\color{light-gray}{MolmoE-A1B-7B}}\\
     & TextVQA~\citep{TextVQA} \newline \tiny{\color{light-gray}{Text Understanding}} & \cellcolor{front-color}50.2\% & \cellcolor{front-color}60.2\% & \cellcolor{front-color} 73.0\% & 61.5\%  \ \tiny{\color{light-gray}{MolmoE-A1B-7B}}\\
     & DocVQA~\citep{DocVQA} \newline \tiny{\color{light-gray}{Document Understanding}} & \cellcolor{front-color}58.3\% & \cellcolor{front-color}70.5\% & \cellcolor{front-color}80.0\% & 77.7\%   \ \tiny{\color{light-gray}{MolmoE-A1B-7B}}\\
     & ScienceQA~\citep{ScienceQA} \newline \tiny{\color{light-gray}{High-school Science}} & \cellcolor{front-color}73.8\% & \cellcolor{front-color}80.0\% & \cellcolor{front-color}89.6\% & 87.5\%   \ \tiny{\color{light-gray}{MolmoE-A1B-7B}} \\
    \midrule
    \multirow{3}{*}{Multi-task} 
     & MMMU~\citep{MMMU} \newline \tiny{\color{light-gray}{College-level Multidiscipline}} & \cellcolor{front-color}29.0\% & \cellcolor{front-color}33.7\% & \cellcolor{front-color}42.0\% & 33.9\%  \ \tiny{\color{light-gray}{MolmoE-A1B-7B}} \\
     & MathVista~\citep{mathvista} \newline \tiny{\color{light-gray}{General Math Understanding}} & \cellcolor{front-color}35.9\% & \cellcolor{front-color}40.1\% & \cellcolor{front-color}51.5\% & 37.6\%  \ \tiny{\color{light-gray}{MolmoE-A1B-7B}} \\
     & MMStar~\citep{MMStar} \newline \tiny{\color{light-gray}{Multidisciplinary Reasoning}} & \cellcolor{front-color}34.6\% & \cellcolor{front-color}38.3\% & \cellcolor{front-color}46.0\% & 43.1\% \ \tiny{\color{light-gray}{MolmoE-A1B-7B}} \\
    \midrule
    \multirow{5}{*}{Video}
     & Video-MME~\citep{video-mme} \newline \tiny{\color{light-gray}{General Video Understanding}} & \cellcolor{front-color}33.7\% & \cellcolor{front-color}42.2\% & \cellcolor{front-color}52.1\% & 45.0\%  \ \tiny{\color{light-gray}{InternVL2-2B}} \\
     & MLVU~\citep{mlvu} \newline \tiny{\color{light-gray}{MovieQA + MSRVTT-Cap}} & \cellcolor{front-color}40.6\% & \cellcolor{front-color}47.3\% & \cellcolor{front-color}55.2\% & 48.2\% \ \tiny{\color{light-gray}{InternVL2-2B}}  \\
     & MVBench~\citep{mvbench} \newline \tiny{\color{light-gray}{Multiview Reasoning}} & \cellcolor{front-color}32.7\% & \cellcolor{front-color}39.7\% & \cellcolor{front-color}46.3\% & 60.2\%  \ \tiny{\color{light-gray}{InternVL2-2B}} \\
     & WorldSense~\citep{worldsense} \newline \tiny{\color{light-gray}{Temporal + Physics}} & \cellcolor{front-color}29.7\% & \cellcolor{front-color}30.6\% & \cellcolor{front-color}36.2\% & 32.4\%  \ \tiny{\color{light-gray}{Qwen2VL-7B}} \\
     & TempCompass~\citep{tempcompass} \newline \tiny{\color{light-gray}{Temporal Understanding}} & \cellcolor{front-color}43.1\% & \cellcolor{front-color}49.0\% & \cellcolor{front-color}53.7\% & 53.4\% \ \tiny{\color{light-gray}{InternVL2-2B}}  \\
    \midrule
    Average & Across Benchmarks & \cellcolor{front-color} 44.0\% & \cellcolor{front-color}51.0\% & \cellcolor{front-color}59.8\% & -- \\
    \midrule
    \multirow{2}{*}{RAM Usage} & Batch size = 1 & \cellcolor{front-color}0.8 GB & \cellcolor{front-color}1.2 GB & \cellcolor{front-color}4.9 GB &  27.7 GB \ \tiny{\color{light-gray}{MolmoE-A1B-7B}} \\
     & batch size = 64 & \cellcolor{front-color}15.0 GB & \cellcolor{front-color}16.0 GB & \cellcolor{front-color}49.9 GB & -- \\
    \bottomrule
    \end{tabular}}
    \caption{
    \textbf{Benchmark comparison of SmolVLM variants across vision-language tasks.}
    Performance of SmolVLM models at three scales (256M, 500M, and 2.2B parameters) compared to efficient open-source models on single-image, multi-task, and video benchmarks. SmolVLM models demonstrate strong accuracy while maintaining significantly lower RAM usage, highlighting their computational efficiency for resource-constrained multimodal scenarios.
    }
    \label{tab:smolvlm_vision_video_size_comp}
\end{table}

%% file: sections/related.tex
\section{Related Work}

\subsection{First-Generation Vision-Language Models}

Early multimodal models achieved significant progress primarily by scaling parameters, but their high computational demands limited practical deployment. For instance, Flamingo~\citep{alayrac2022flamingo}, an $80$B-parameter Vision-Language Model (VLM), integrated a frozen $70$B-parameter LM~\citep{hoffmann2022training} with a vision encoder employing gated cross-attention and a Perceiver Resampler~\citep{perceiver} for efficient token compression. Despite state-of-the-art few-shot capabilities without task-specific fine-tuning, Flamingo's large scale posed significant deployment challenges.

Hugging Face’s Idefics~\citep{OBELICS} adopted Flamingo’s architecture, offering models at both 9B and 80B parameters, further exemplifying the approach of large-scale multimodal training. In contrast, BLIP-2~\citep{BLIP-2} proposed a more parameter-efficient, modular design by freezing both the vision encoder and language model, introducing instead a lightweight Query Transformer (Q-Former) that translates visual features into language-compatible tokens. This approach significantly reduced trainable parameters, surpassing Flamingo’s performance on VQA tasks~\citep{antol2015vqa,goyal2017making} with roughly 54 times fewer trainable parameters, thus paving the way toward more efficient multimodal architectures.

Similarly, LLaVA (Large Language-and-Vision Assistant)~\citep{LLaVA} connected a pretrained CLIP~\citep{radford2021learning} ViT image encoder to a LLaMA/Vicuna language backbone~\citep{touvron2023llama,Vicuna}, fine-tuning the combined model on instruction-following datasets. Resulting in a 13B-parameter multimodal chatbot with GPT-4V-like capabilities~\citep{GPT4}, LLaVA achieved notable visual conversational performance. However, despite being smaller and faster than Flamingo, it still demands substantial GPU memory for real-time interaction and inherits the limitations of the underlying language model’s context window (typically 2048 tokens).

Recent research has actively explored various design choices, training strategies, and data configurations to enhance Vision-Language Models (VLMs). For instance, Idefics2~\citep{Idefics2} introduced architectural and training-data improvements compared to its predecessor, advancing open-source VLM capabilities. Concurrently, Cambrian1~\citep{tong2024cambrian} examined fundamental design principles and scaling behaviors, aiming for more efficient architectures. Projects like Eagle~\citep{shi2024eagle} and its successor Eagle2~\citep{li2025eagle} have optimized specific architectural components, targeting improved performance and efficiency. Additionally, recent efforts such as Apollo~\citep{zohar2024apollo} extend multimodal architectures from static images to video understanding, further enriching the diversity of approaches.

\subsection{Efficiency-Focused Vision-Language Models}

Larger models, such as InternVL \citep{chen2023internvl, chen2024internvl25} and Qwen-VL \citep{bai2023qwenvlversatilevisionlanguagemodel, bai2025qwen25vl, wang2024qwen2}, introduced architectural innovations for improved computational efficiency. InternVL aligns a 6B-parameter vision transformer (ViT) with an 8B-parameter language "middleware," forming a 14B-parameter model that achieves state-of-the-art results across multiple vision and multimodal tasks. This balanced architecture narrows the modality gap, enabling robust multimodal perception and generation capabilities. Similarly, Qwen-VL integrates a Qwen language model with specialized visual modules, leveraging captioned bounding-box data to enhance visual grounding and text recognition capabilities. Despite its strong multilingual and multimodal performance, Qwen-VL generates exceptionally long token sequences for high-resolution inputs, increasing memory requirements.

On the smaller end, models like PaliGemma, Moondream2, and MiniCPM-V demonstrate impressive multimodal capabilities within constrained parameter budgets. PaliGemma \citep{team2024gemma}, with just 3B parameters (400M vision encoder from SigLIP-So \citep{li2023siglip} and 2B Gemma language model), effectively covers a wide range of multimodal tasks. However, its condensed visual interface can limit detailed visual analysis. Moondream2, at merely 1.8B parameters, pairs SigLIP visual features with Microsoft's Phi-1.5 language model \citep{li2023textbooks}, showcasing competitive performance on tasks such as image description, OCR, counting, and classification, ideal for edge and mobile applications. MiniCPM-V \citep{hu2024minicpmunveilingpotentialsmall}, specifically designed for on-device scenarios, integrates a 400M vision encoder and a 7.5B language model via a perceiver-style adapter. This compact model notably achieves GPT-4V-level performance on selected benchmarks.
 Deepseek VL and Deepseek VL2 \citep{DeepSeek-VL, DeepSeek-VL2}, spanning 2–7B and 4–27B parameters respectively, further illustrate the growing focus on efficient yet powerful multimodal models suitable for resource-constrained environments. Collectively, these models demonstrate the increasing feasibility of deploying effective, real-time multimodal AI in practical scenarios.

\subsection{Multimodal Tokenization and Compression Strategies}

Efficient tokenization significantly reduces computational and memory demands in Vision-Language Models (VLMs). Early methods, encoding every pixel or patch individually, resulted in lengthy sequences—196 tokens for a 224$\times$224 image at 16$\times$16 resolution. Recent strategies compress visual data while preserving essential details.
Learned modules like Perceiver Resamplers \citep{perceiver} used by Flamingo and Idefics2 \citep{alayrac2022flamingo,Idefics2}, and BLIP-2’s Q-Former \citep{BLIP-2}, compress inputs into a small set of latent tokens. While effective in shortening sequences, these methods may limit performance on fine-grained tasks like OCR \citep{TextVQA,STVQA}.
Spatial compression via patch pooling and pixel shuffle is increasingly popular. InternVL v1.5 and Idefics3 \citep{chen2023internvl,chen2024internvl25,OBELICS} use 2$\times$2 pixel-shuffle, reducing token counts fourfold while maintaining OCR capability.
Models like Qwen-VL-2 \citep{wang2024qwen2} adopt multi-scale representations and selective token dropping via convolutional and Transformer modules. Adaptive methods, such as image tiling in UReader and DocOwl, dynamically adjust token counts based on task complexity, sacrificing some global context.

\subsection{Video-Capable Vision-Language Models}

Extending vision-language models (VLMs) from images to videos significantly increases complexity due to temporal dimensions, expanding token counts and computational demands. Early models, such as Video-LLaVA \citep{lin2023video}, unified image and video training, aligning video frame features with static images and substantially outperforming predecessors like Video-ChatGPT \citep{maaz2023video} on benchmarks including MSRVTT \citep{xu2016msr}, MSVD \citep{chen2011collecting}, TGIF \citep{li2016tgif}, and ActivityNet \citep{caba2015activitynet}. Meanwhile, Video-STaR~\citep{video-star} introduced the first self-training approach that leverages existing labeled video datasets for instruction tuning of Large Multimodal Models.

Recent models enhance efficiency and effectiveness in handling long-form video content. Temporal Preference Optimization (TPO) \citep{li2025temporal} employs self-training with localized and comprehensive temporal grounding, improving benchmarks like LongVideoBench, MLVU, and Video-MME. Oryx MLLM \citep{liu2024oryx} dynamically compresses visual tokens via its OryxViT encoder, balancing efficiency and precision across tasks. 
VideoAgent \citep{wang2024videoagent} models long-form video understanding as a decision-making process, utilizing a large language model (LLM) as an agent to identify and compile crucial information for question answering iteratively. 
VideoLLaMA3 \citep{zhang2025videollama} adapts its vision encoder for variable resolutions and uses multi-task fine-tuning to enhance video comprehension. Video-XL \citep{shu2024video} introduces Visual Summarization Tokens (VST) and curriculum learning for efficient handling of hour-scale videos. Similarly, Kangaroo \citep{liu2024kangaroo} utilizes curriculum training to scale input resolution and frame count progressively, achieving top performance on diverse benchmarks.

Apollo \citep{zohar2024apollo} recently made an in-depth exploration of Video-LMMs and showed the architecture and training schedule that most affect performance. In so doing, it showed the remarkable efficiency gains that can be made during training and inference. Apollo achieved state-of-the-art results with modest parameter sizes on benchmarks such as LongVideoBench, MLVU, and Video-MME \citep{mlvu,video-mme}.